\documentclass[conference]{IEEEtran}

\IEEEoverridecommandlockouts

\usepackage{cite}
\usepackage{amsmath,amssymb,amsfonts}
\usepackage{algorithmic}
\usepackage{graphicx}
\usepackage{textcomp}
\usepackage{xcolor}
\usepackage{multirow}
\usepackage{makecell}
\usepackage{array}
\usepackage{balance}
\usepackage{subcaption}

\usepackage{pgfplots}
\usepgfplotslibrary{external}
\def\BibTeX{{\rm B\kern-.05em{\sc i\kern-.025em b}\kern-.08em
    T\kern-.1667em\lower.7ex\hbox{E}\kern-.125emX}}
\begin{document}
\bstctlcite{BSTcontrol}

\title{Hierarchical Proportion Models for Motion Generation via Integration of Motion Primitives
\thanks{This work was supported by JSPS KAKENHI Grant Number 24K00905, JST, PRESTO Grant Number JPMJPR24T3 Japan, and JST ALCA-Next Japan, Grant Number JPMJAN24F1. This study was based on the results obtained from the JPNP20004 project subsidized by the New Energy and Industrial Technology Development Organization (NEDO).}
}

\author{\IEEEauthorblockN{1\textsuperscript{st} Yu-Han, Shu}
\IEEEauthorblockA{\textit{Intelligent and Mechanical Interaction Systems} \\
\textit{University of Tsukuba}\\
Tsukuba, Japan \\
shu.yuhan.tkb\_en@u.tsukuba.ac.jp}
\and
\IEEEauthorblockN{2\textsuperscript{nd} Toshiaki Tsuji}
\IEEEauthorblockA{\textit{Science and Engineering} \\
\textit{Saitama University}\\
Saitama, Japan \\
tsuji@ees.saitama-u.ac.jp}
\and
\IEEEauthorblockN{3\textsuperscript{rd} Sho Sakaino}
\IEEEauthorblockA{\textit{Systems and Information Engineering} \\
\textit{University of Tsukuba}\\
Tsukuba, Japan \\
sakaino@iit.tsukuba.ac.jp}
}

\maketitle
\begin{abstract}
Imitation learning (IL) enables robots to acquire human-like motion skills from demonstrations, but it still requires extensive high-quality data and retraining to handle complex or long-horizon tasks. To improve data efficiency and adaptability, this study proposes a hierarchical IL framework that integrates motion primitives with proportion-based motion synthesis. The proposed method employs a two-layer architecture, where the upper layer performs long-term planning, while a set of lower-layer models learn individual motion primitives, which are combined according to specific proportions. Three model variants are introduced to explore different trade-offs between learning flexibility, computational cost, and adaptability: a learning-based proportion model, a sampling-based proportion model, and a playback-based proportion model, which differ in how the proportions are determined and whether the upper layer is trainable. Through real-robot pick-and-place experiments, the proposed models successfully generated complex motions not included in the primitive set. The sampling-based and playback-based proportion models achieved more stable and adaptable motion generation than the standard hierarchical model, demonstrating the effectiveness of proportion-based motion integration for practical robot learning.
\end{abstract}

\begin{IEEEkeywords}
Imitation Learning, Bilateral Control, Model Predictive Control, Motion Planning, Intelligent Robotics.
\end{IEEEkeywords}

\section{Introduction}
Robotic technologies that support household, human support, and production activities are increasingly anticipated. In recent years, approaches based on machine learning have been actively studied to enable robotic behaviors that adapt to changing environments \cite{Wang2021}. Among these, imitation learning (IL) has emerged as a promising method to efficiently transfer human motion skills to robots \cite{Zare2023}. By leveraging demonstrations, IL improves data efficiency and reduces training time.

In recent years, research on IL utilizing force information and bilateral control has further enhanced flexibility and stability in robotic motions \cite{Tsuji2025, Adachi2018}. Bilateral control is a teleoperation technique between robots, which involves the synchronization of position and force. Compared to conventional demonstration methods, such as direct teaching or virtual reality-based teaching \cite{Zhang2018}, learning from data collected through bilateral control has achieved faster motion while maintaining equivalent flexibility \cite{Sasagawa2021, Kato2021}.

Nevertheless, IL still requires large volumes of high-quality demonstrations, which is especially challenging for long-horizon or complex tasks \cite{Zheng2024}. Repeated data collection and retraining are necessary to achieve task success or adapt to new tasks, which significantly limits its practical deployment.

A potential solution is to decompose tasks into reusable motion primitives that can be recombined to generate novel tasks \cite{Gao2024}. Previous studies following this concept have proposed approaches in which individual motion primitives are learned by expert models, while a higher-level controller selects the optimal model at each time step \cite{Ito2022, Luo2024}.

The Mixture of Experts (MoE) framework offers a promising alternative by integrating multiple expert outputs through weighted averaging, enabling smooth transitions and adaptability \cite{Cai2024}. In the context of robotic motion generation, MoE has also demonstrated promising results in simulation; however, verification in real-world environments remains limited \cite{Celik2021}. 

In parallel, Monte Carlo Model Predictive Control (MC-MPC) provides sampling-based optimization that generates diverse motion candidates and selects the optimal one based on evaluation values, achieving precise and robust performance under uncertainty \cite{Janson2017}. This capability makes MC-MPC particularly effective for handling unknown environments and deformable object manipulation \cite{Nakamura2024}.

Building on these prior works, this study proposes an extended approach that integrates a hierarchical structure into IL based on bilateral control \cite{Hayashi2022}. The method employs a two-layer model: the lower layer learns individual motion primitives, while the upper layer performs long-term planning. Afterward, the task motions are generated by computing a weighted average of the motion primitives according to specific proportions. Within this framework, three model variants are investigated, which differ in how the combination proportions are determined and whether the upper layer is learned:
\begin{enumerate}
  \item Learning-based proportion model: The upper layer learns the long-term planning and the proportions for combining motion primitives. This model was previously introduced in our earlier study \cite{Shu2025}.
  \item Sampling-based proportion model: The upper layer learns only the long-term planning. The proportions are derived from the difference between upper- and lower-layer outputs.
  \item Playback-based proportion model: The upper layer is replaced with motion data, and the proportions are computed from the difference between data and lower-layer outputs.
\end{enumerate}

Compared with the original hierarchical model, which requires task-specific training of both layers, the proposed framework emphasizes reusable motion primitives and task adaptation through proportion determination. In particular, the learning-based proportion model learns the combination proportions and typically requires retraining when the primitive set is modified. In contrast, the sampling-based and playback-based variants determine proportions without proportion-level learning, enabling more stable motion generation and simpler updates of the lower-layer models.

The effectiveness of the proposed models was validated through pick-and-place experiments, including a complex task not included in the primitive set. While the learning-based proportion model faced challenges in estimating appropriate proportions when many primitives were involved, the sampling-based and playback-based proportion models achieved higher motion quality. Moreover, the lower-layer models can be shared across multiple tasks, and the playback-based proportion model can even operate without upper-layer learning, thereby reducing the overall learning cost.

This paper is organized as follows: Section \ref{sec2} presents the fundamental background. Section \ref{sec3} describes the proposed model, detailing its design and implementation. Section \ref{sec4} reports the experimental setup and results. Finally, Section \ref{sec5} concludes the study and outlines potential directions for future research.

\section{Technical Foundations}
\label{sec2}
\subsection{Bilateral Control}
Bilateral control is a teleoperation technique that employs two robots: a leader and a follower. The leader is directly manipulated by an operator, while the follower reproduces the leader's motion. During operation, the two robots synchronize their joint angles $\boldsymbol{\theta}$ and angular velocities $\boldsymbol{\omega}$, while exchanging torque feedback $\boldsymbol{\tau}$ with each other. The control objectives of the system are defined as follows:
\begin{align}
    \boldsymbol{\theta}^{res}_l-\boldsymbol{\theta}^{res}_f=0,\\
    \boldsymbol{\tau}^{res}_l+\boldsymbol{\tau}^{res}_f=0,
    \label{eq:bilateralControl}
\end{align}
where the superscripts $l$ and $f$ denote the leader and follower states, respectively, and the subscript $res$ represents the response values. Bilateral control has been shown to enable fast and stable motion by providing synchronized position and force feedback between the two robots \cite{Sasagawa2020}.

\subsection{Hierarchical Model}
Previous studies have demonstrated that when the predicted response of the leader obtained through bilateral control is used as the command input for the follower, the robot can reproduce appropriate imitative motions~\cite{Adachi2018}. In this approach, the model outputs the next leader prediction $\hat{L}_{k+1}$ directly from the follower's response $F_k$. To further enhance the efficiency of long-term motion generation in bilateral control-based imitation learning, a hierarchical model was proposed~\cite{Hayashi2022}.

A hierarchical model processes information across multiple layers, enabling robots to perform long-term tasks more effectively through IL. However, conventional single Long Short-Term Memory (LSTM) networks have limited memory capacity, which reduces efficiency and prediction accuracy for longer sequences. The hierarchical model addresses this issue by dividing responsibilities between two neural networks: the upper layer manages long-term task planning, while the lower layer captures short-term detailed motions. Based on the long-term prediction provided by the upper layer, the lower layer constructs command samples for the next step, resulting in more adaptive and consistent motion generation.

\begin{figure}[t]
    \centering
    \includegraphics[width=0.65\linewidth]{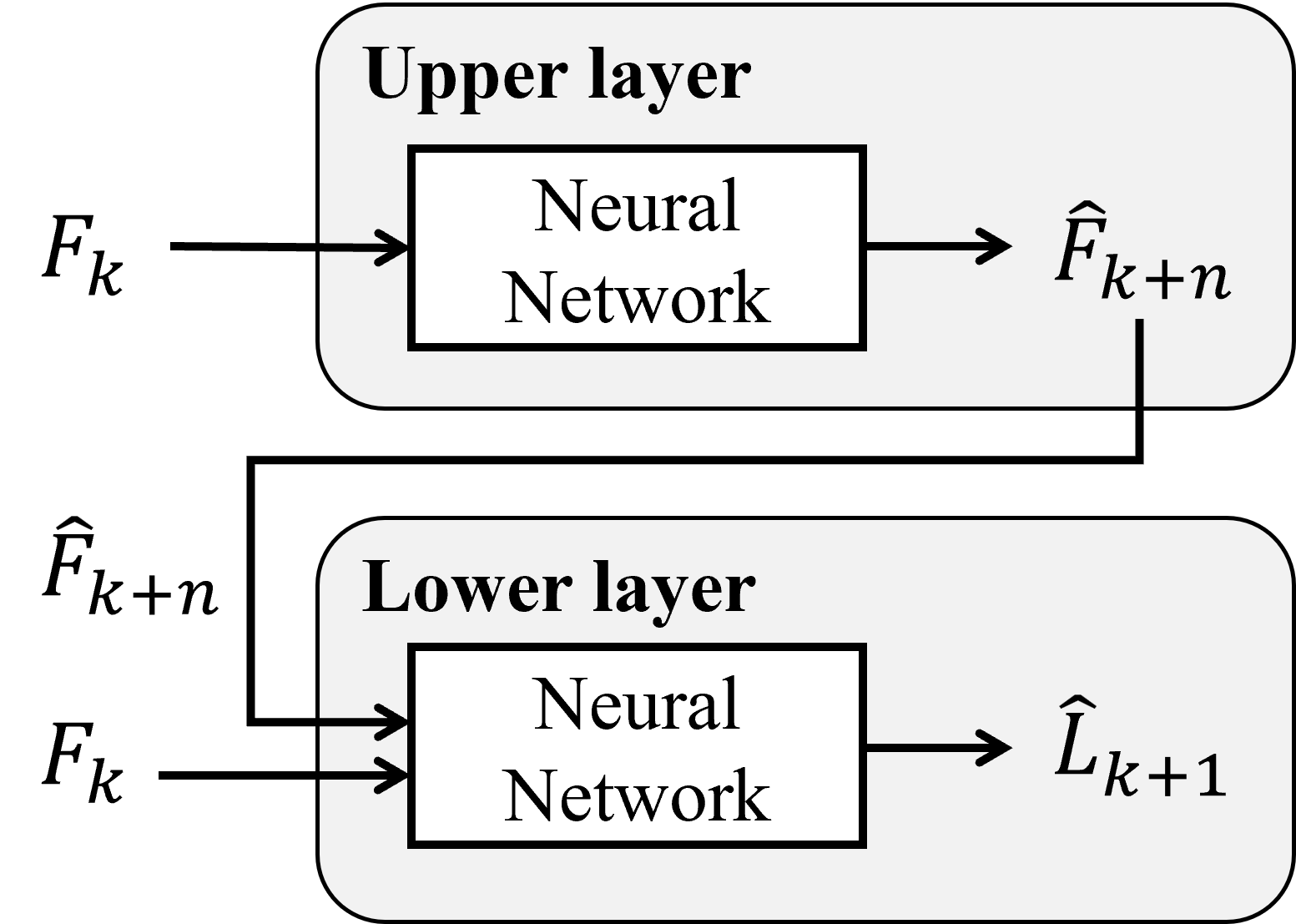}
    \caption{Structure of hierarchical model.}
    \label{fig:HierarchicalModel}
\end{figure}

The structure of the hierarchical model is illustrated in Figure~\ref{fig:HierarchicalModel}, where $\boldsymbol{F}$ and $\boldsymbol{L}$ denote the state vectors $[\boldsymbol{\theta}, \boldsymbol{\omega}, \boldsymbol{\tau}]$ of the follower and leader, respectively. Within this framework, the upper layer predicts the follower state $\hat{F}_{k+n}$ $n$ steps ahead based on the current follower state $F_k$, with updates occurring every $n$ time steps. Meanwhile, the lower layer predicts the next leader state $\hat{L}_{k+1}$ at each step using the upper-layer output $\hat{F}_{k+n}$ and the current follower state $F_k$.
%
\subsection{Monte Carlo Model Predictive Control (MC-MPC)}
MC-MPC is a sampling-based control method that determines optimal input sequences by generating a large set of candidates and simulating their outcomes over a prediction horizon~\cite{Nakatani2019}. In contrast to traditional MPC, MC-MPC does not depend on smooth cost functions or accurate system dynamics, making it appropriate to highly nonlinear and discontinuous problems such as collisions or slippage. Moreover, its compatibility with parallel computation enables efficient real-time performance, making MC-MPC a robust, flexible, and practical framework for robot motion generation in complex environments.

\begin{figure}[t]
    \centering
    \includegraphics[width=.95\linewidth]{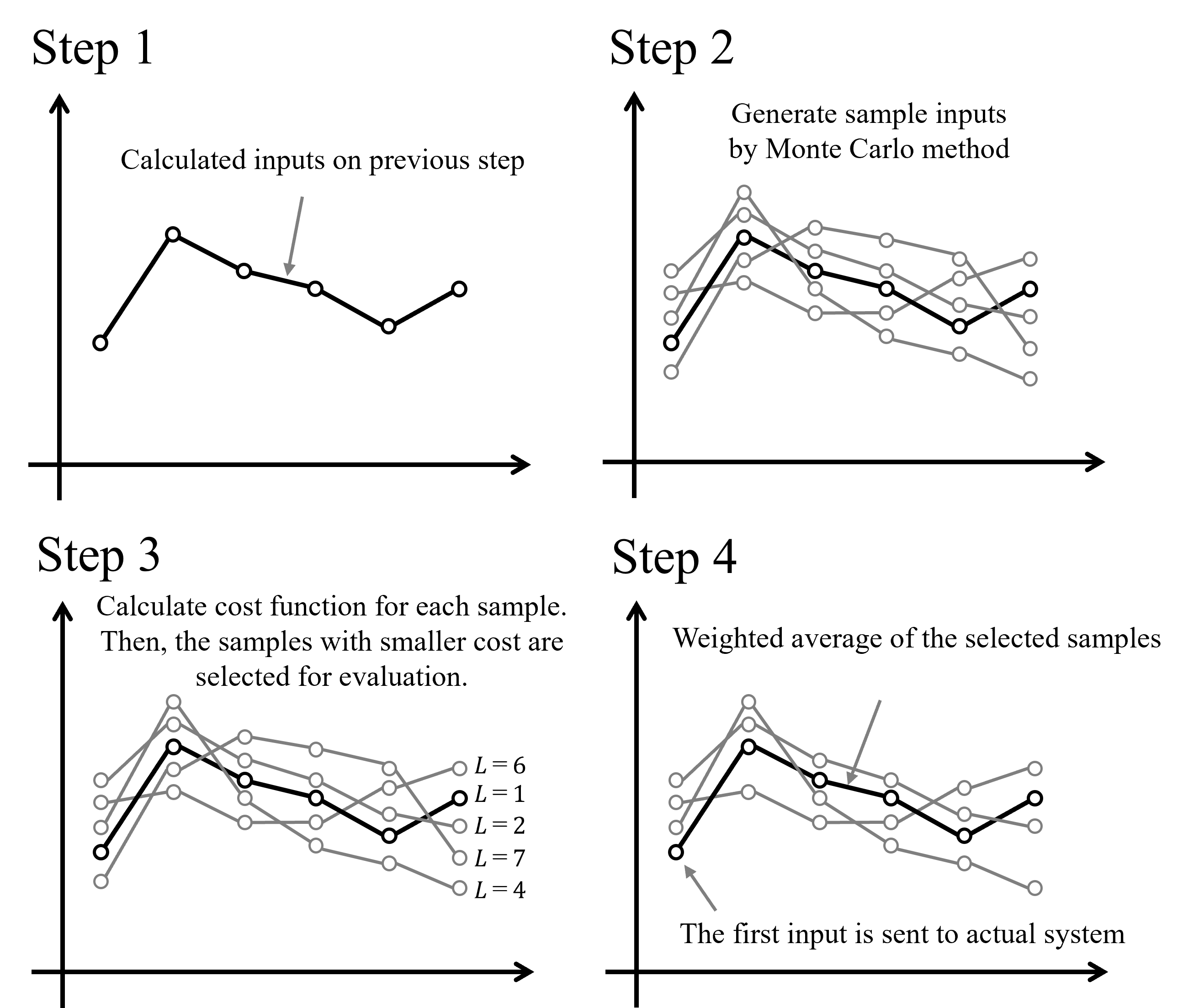}
    \caption{Algorithm of MC-MPC.}
    \label{fig:MCMPC}
\end{figure}

As illustrated in Figure~\ref{fig:MCMPC}, the algorithm begins with an initial trajectory derived from the previous cycle's optimal sequence. Multiple randomized input sequences are then generated in parallel to explore diverse control strategies. Each sample is evaluated using a cost function, and the top-performing candidates are selected to compute a weighted average as the final input sequence. Subsequently, the first element of the optimized input sequence is applied to the system, while the remaining elements are retained for the next control cycle. The weights for the averaging process are computed through a cross-entropy formulation, expressed as follows:
\begin{align}
\boldsymbol{u}_{{k|i}} = \frac{\sum_{m\in M}\boldsymbol{u}^m_{{k|i}} \exp\left(-\frac{L^m}{\rho}\right)}{\sum_{m\in M} \exp\left(-\frac{L^m}{\rho}\right)}
\label{eq:cross_entropy}
\end{align}
where $\boldsymbol{u}^m_{k|i}$ denotes the input at step $i$ of sample $m$ in control cycle $k$, $L^m$ is the corresponding cost function value, $\rho$ is the temperature parameter, and $M$ represents the set of top-performing samples.
%
\section{Proposed Method}
\label{sec3}
In this section, we present the proposed hierarchical model, in which the lower-layer model learns simpler primitive motions decomposed from a complex task. The desired task motion is then constructed as a weighted average of these primitives, with the weighting scheme varying across the different proposed models.

\subsection{Decomposition and Combination of Complex Motions}
\begin{figure}[t]
    \centering
    \includegraphics[width=.95\linewidth]{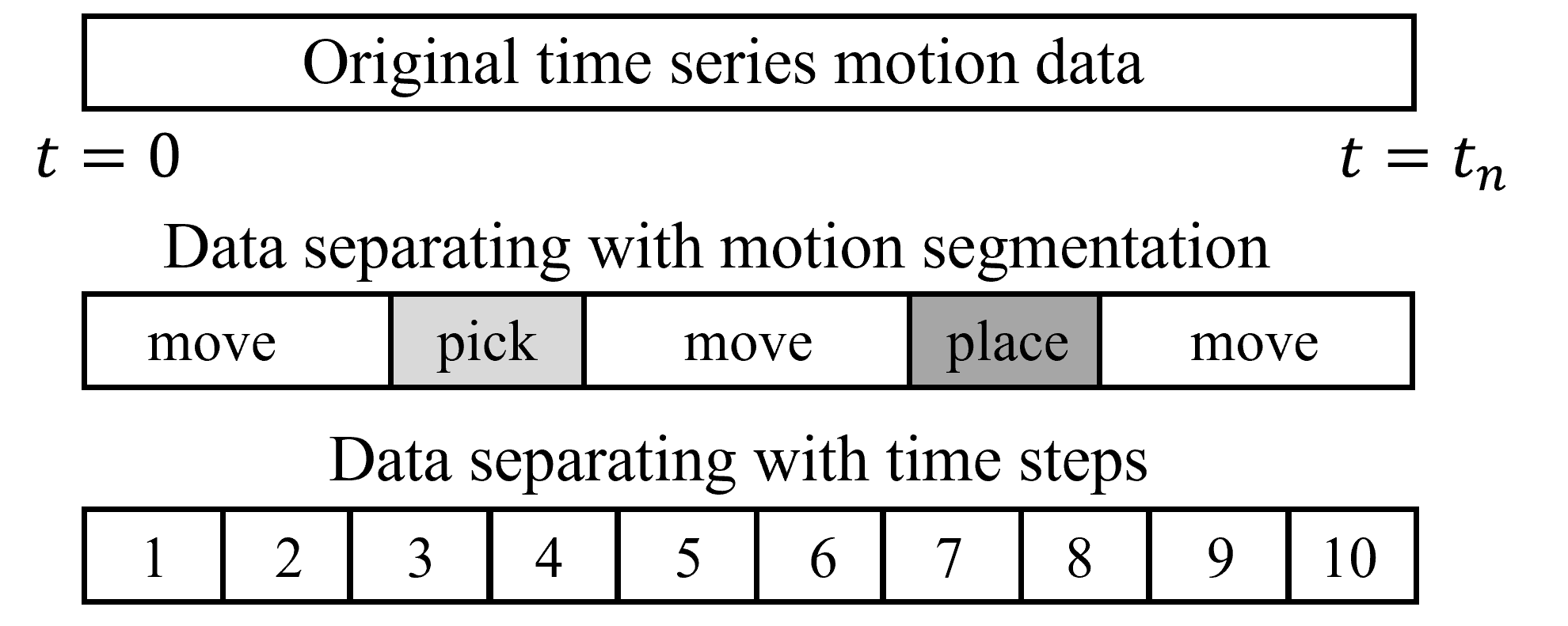}
    \caption{Comparison between different data separation methods.}
    \label{fig:dataSeparation}
\end{figure}
A natural approach to decomposing complex tasks is to separate them based on action contents. For example, in the pick-and-place task, the robot grasps an object, moves and places it at the destination, which can be divided into “grasping,” “moving,” and “placing”. With this decomposition strategy, the lower-layer models learn each decomposed motion primitive, while an upper-layer model combines them to generate the overall task \cite{Luo2024}.

In this study, we introduce a motion decomposition strategy without task-specific segmentation to enhance generalization.
As shown in Figure~\ref{fig:dataSeparation}, motion primitives are separated using uniform time-based separation instead.
By combining these primitives, the model is expected to achieve greater flexibility, enabling the execution of a broader range of tasks, including those not explicitly learned. For instance, a new motion such as “rotating” may be generated by appropriately combining the primitives “grasping,” “transporting,” and “placing.”
\subsection{Proposed Model}
We propose three models that differ in the method used to determine the combination proportions of motion primitives and in whether the upper layer is implemented as a trained network or as target data.

\subsubsection{Learning-based Proportion Model}
\begin{figure}[t]
    \centering
    \includegraphics[width=.9\linewidth]{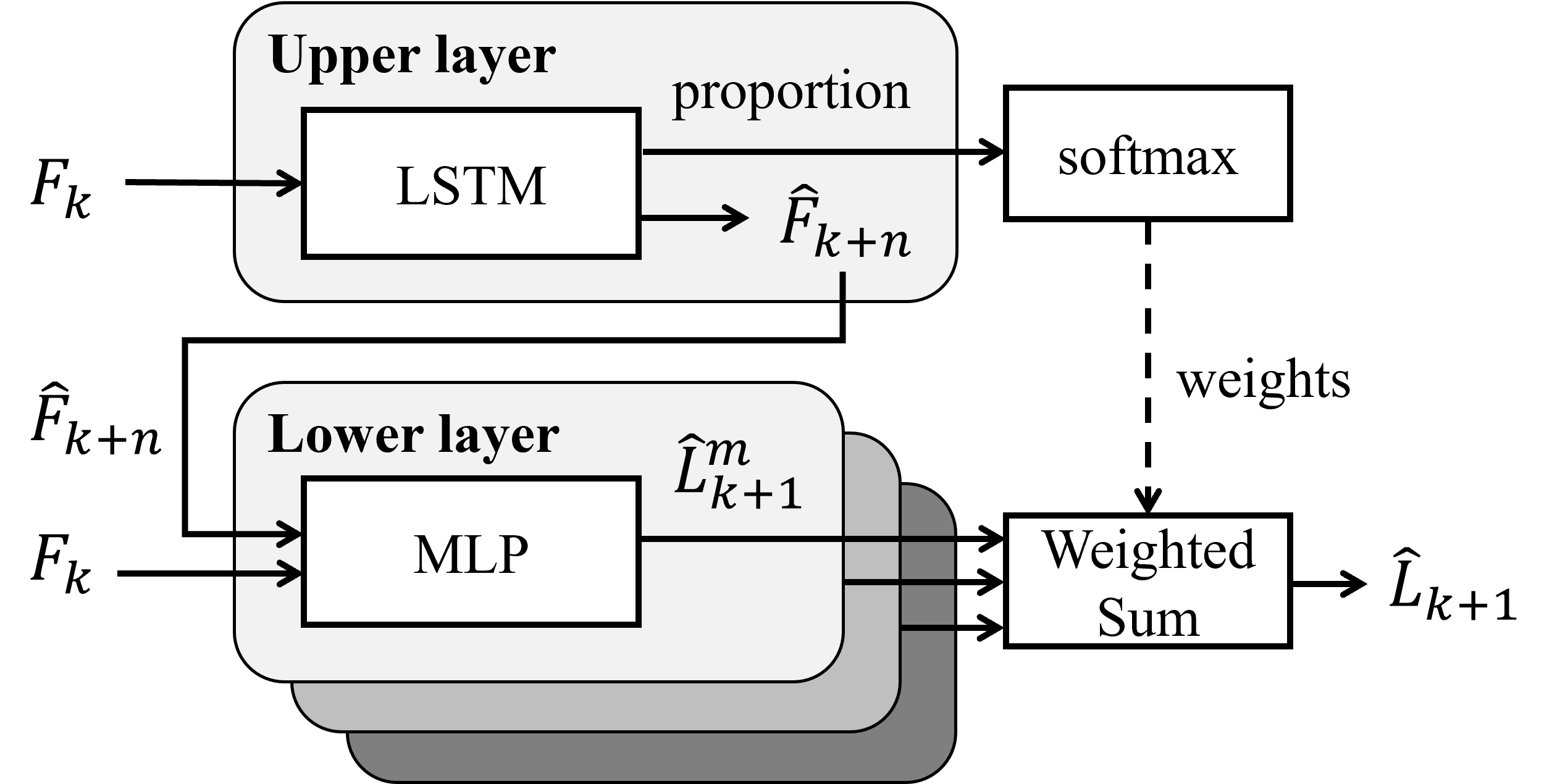}
    \caption{Structure of the learning-based proportion model.}
    \label{fig:model}
\end{figure}
The structure of the proposed hierarchical model is illustrated in Figure~\ref{fig:model}. The model consists of multiple lower-layer models that generate specific motion primitives, and an upper layer that learns both the combination proportions of these primitives and long-term motion planning. To ensure that the generated actions remain within a reasonable range, the output proportions are passed through a softmax function. Based on these proportions, a weighted average of the lower-layer outputs is calculated to produce the execution command for the next time step.

To capture long-term dependencies across time steps in motion data, an LSTM network is employed in the upper layer. In contrast, since long-term information processing is unnecessary in the lower layer, each lower-layer model is implemented as a multi-layer perceptron (MLP) without internal states.

The training procedure begins with the training of the lower-layer models for each motion primitive. Afterward, the process shifts to the upper-layer model, where the pre-trained lower-layer models are executed together with the upper layer to learn the optimal proportions of the motion primitives. A softmax function is applied to the proportion outputs to ensure that the combined control input remains within a valid joint position range.

\subsubsection{Sampling-based Proportion Model}
\begin{figure*}[t]
    \centering
    \includegraphics[width=.9\linewidth]{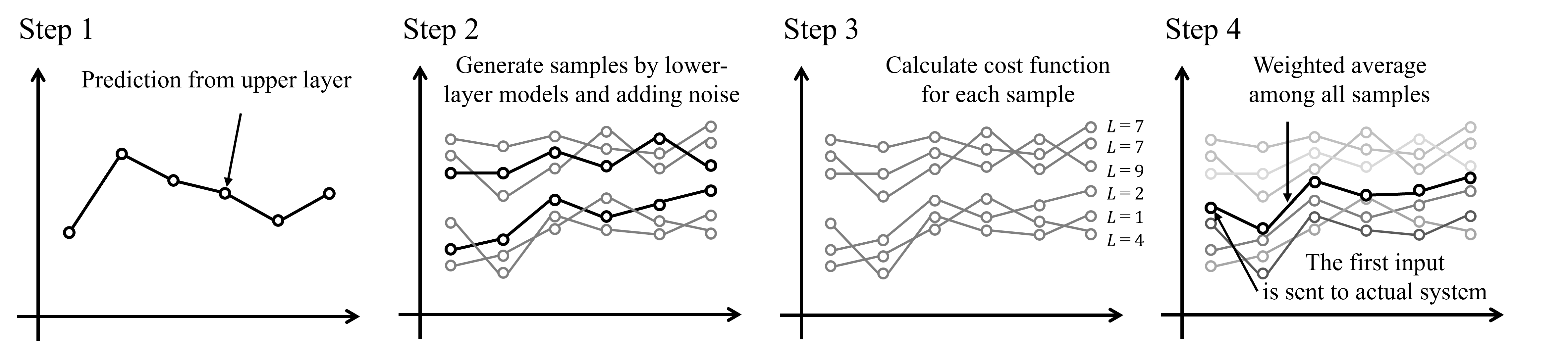}
    \caption{Algorithm of proposed sampling-based proportion model.}
    \label{fig:MCMPCModel_concept}
\end{figure*}
\begin{figure}[!t]
    \centering
    \includegraphics[width=1.\linewidth]{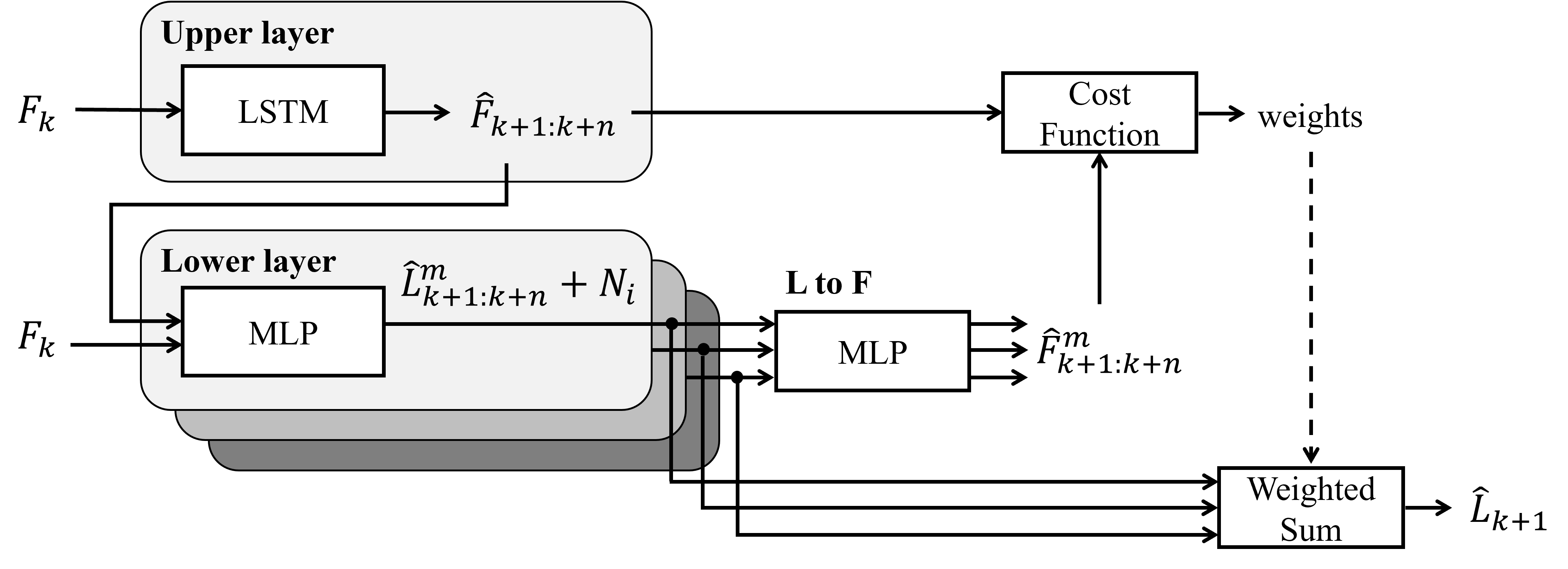}
    \caption{Structure of the proposed sampling-based proportion model.}
    \label{fig:MCMPCModel}
\end{figure}

This section introduces a framework that integrates the concept of MC-MPC into the hierarchical structure. Similar to MC-MPC, the basic idea is to generate a large number of randomized samples, evaluate them, and compute their weighted average. Specifically, the upper layer predicts future follower states based on the current system state. Using this prediction, the lower layer generates leader trajectories with added noise to increase sample diversity. Each sample is evaluated using a cost function, and the weighted average of the top-performing samples is used to produce the next command. The conceptual overview is illustrated in Figure~\ref{fig:MCMPCModel_concept}.

The overall architecture is presented in Figure~\ref{fig:MCMPCModel}. The upper layer outputs a sequence of long-term follower states, $\hat{\boldsymbol{F}}_{k+1:k+n}$. Based on this, the lower layer predicts leader states, $\hat{\boldsymbol{L}}_{k+1:k+n}^m$, and noise $N_i$ is added to generate diverse samples. To compute the cost, the leader outputs are converted into follower commands using a pre-trained mapping network $\boldsymbol{L}$ to $\boldsymbol{F}$, which is trained on motion data collected through bilateral control. The cost function is defined as follows:
\begin{align*}
\mathcal{L}_\theta &= MSE(\hat{\boldsymbol{\theta}}^m_{k+1} - \hat{\boldsymbol{\theta}}_{k+1}),\\
\mathcal{L}_\omega &= MSE(\hat{\boldsymbol{\omega}}^m_{k+1} - \hat{\boldsymbol{\omega}}_{k+1}),\\
\mathcal{L}_\tau &= MSE(\hat{\boldsymbol{\tau}}^m_{k+1} - \hat{\boldsymbol{\tau}}_{k+1}),\\
\mathcal{L} &= \alpha \cdot \mathcal{L}_\theta + \beta \cdot \mathcal{L}_\omega + \gamma \cdot \mathcal{L}_\tau,
\end{align*}
where $\mathcal{L}_\theta$, $\mathcal{L}_\omega$, and $\mathcal{L}_\tau$ correspond to the joint angle, angular velocity, and torque costs, respectively. The total cost $\mathcal{L}$ is a weighted sum of these terms with coefficients $\alpha$, $\beta$, and $\gamma$. The cross-entropy method~\eqref{eq:cross_entropy} is then applied, assigning higher weights to samples with smaller errors. Using these weights, a weighted average is computed, and the first element $\hat{L}_{k+1}$ is applied to the system, and the above process is repeated.

The key advantage of this architecture compared with the learning-based proportion model is the ease of modifying the lower-layer set. As a result, the number of motion primitives and samples can be readily increased, thereby enhancing motion accuracy and generalization capability.

The training procedures for the upper and lower layers are independent and can be executed in parallel, reducing the overall training time. The L to F model is also trained separately.

\subsubsection{Playback-based Proportion Model}
\begin{figure}[t]
    \centering
    \includegraphics[width=1.\linewidth]{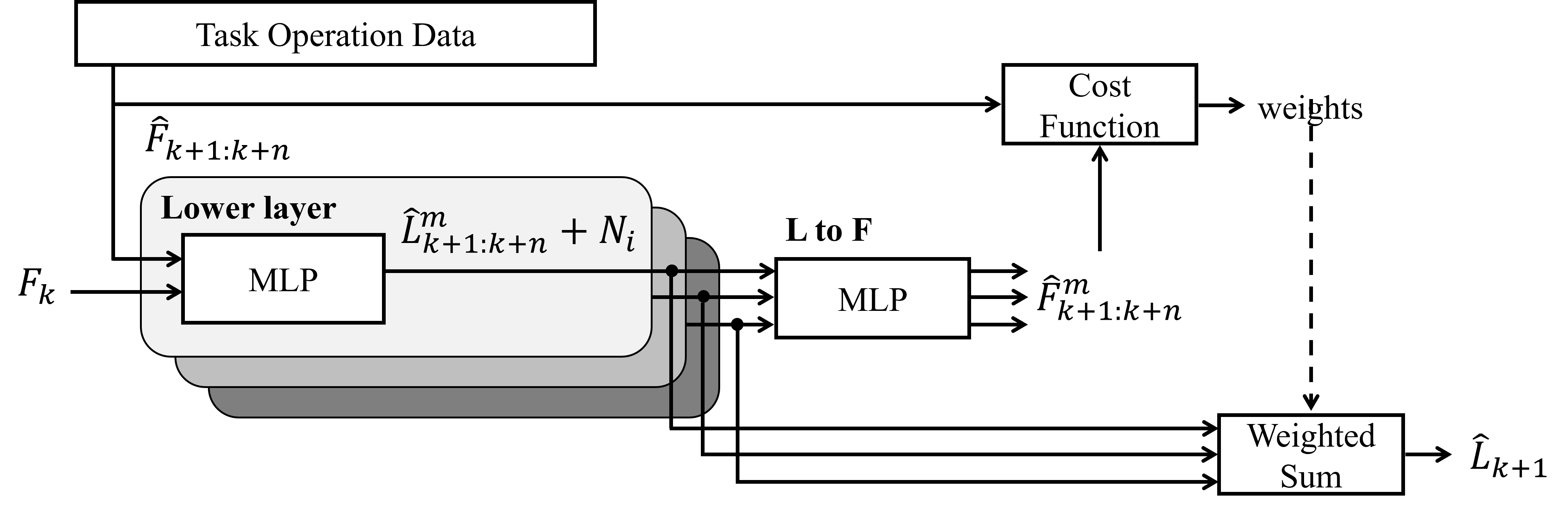}
    \caption{Structure of the proposed playback-based proportion model.}
    \label{fig:model_upperData}
\end{figure}

The model illustrated in Figure~\ref{fig:model_upperData} represents a variation of the sampling-based proportion model, in which the upper layer is replaced with pre-collected motion data. The data required for the lower-layer prediction and cost computation are directly provided by the upper-layer motion data. Similar to the sampling-based proportion model, the lower layer generates multiple samples, which are evaluated against the upper-layer data, and the final outputs are synthesized using the weights derived from equation~\eqref{eq:cross_entropy}.

The key advantage of this configuration is that retraining of the upper layer is unnecessary, allowing rapid adaptation to new tasks or modifications. This flexibility further reduces overall learning costs and enhances responsiveness to environmental and task variations. Although pre-recorded motion data are used, the framework retains key advantages of learning-based characteristics, as the lower-layer models implicitly adapt to variations in execution conditions without explicit retuning of the control parameters.

The training procedure for the lower-layer models is the same as in the other two models, where each learns a corresponding motion primitive. A demonstration dataset of the desired motion is applied to the upper layer.

\section{Experiments}
\label{sec4}
\subsection{Data collection}
\begin{figure}[t]
    \centering
    \begin{subfigure}{.48\linewidth}
        \centering
        \includegraphics[width=1.\linewidth]{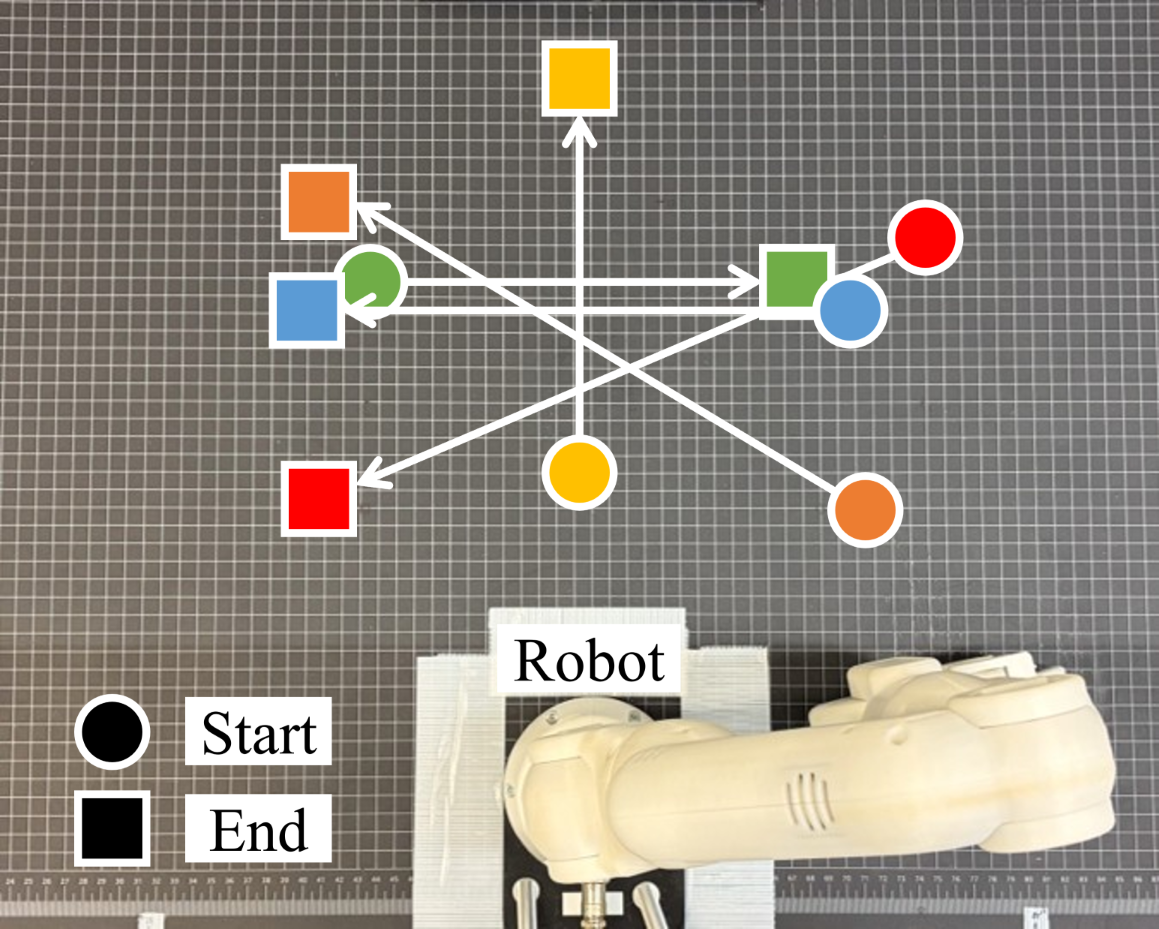}
        \caption{Motion primitives.}
        \label{fig:primitiveSetting}
    \end{subfigure}
    \begin{subfigure}{.48\linewidth}
        \centering
        \includegraphics[width=1.\linewidth]{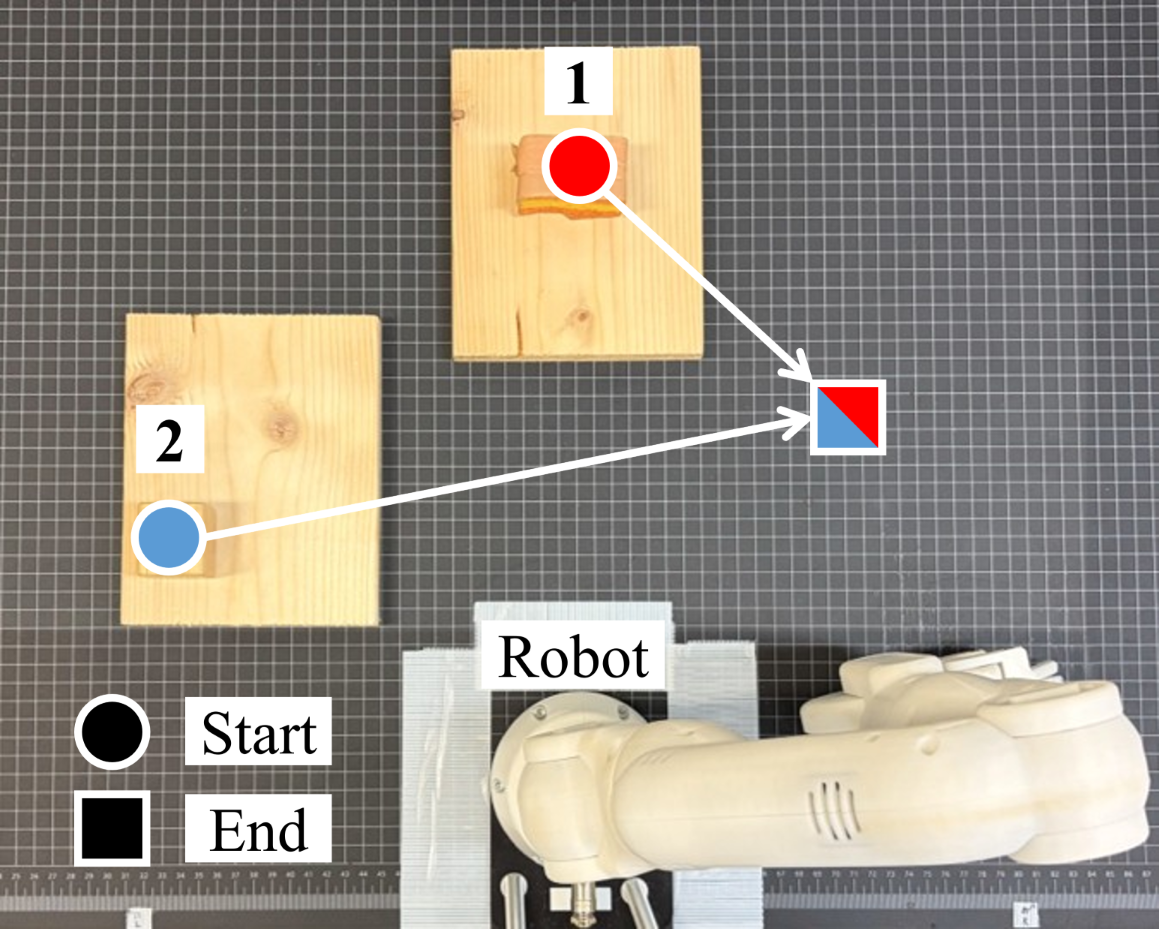}
        \caption{Validation task.}
        \label{fig:taskSetting}
    \end{subfigure}
    \caption{Overview of the experimental environment.}
    \label{fig:Setup}
\end{figure}

A pick-and-place task was conducted to validate the proposed models. In this task, the robot starts from an initial position, moves to the target object, picks it up, and transfers it to a terminal position. For the training data of motion primitives, datasets were prepared for 5 different positions and directions: left-to-right, right-to-left, front-to-back, right-bottom-to-left-upper, and right-upper-to-left-bottom, as illustrated in Figure~\ref{fig:Setup}(\subref{fig:primitiveSetting}). The data were then divided into 10 overlapping segments along the timeline, each ranging from 0.9 to 1.1 times the length of $1/10$ of the total duration. Consequently, 50 motion primitives were used to train the lower-layer models.

For the upper-layer training, two tasks were designed to evaluate the model performance. The first corresponds to the same right-to-left motion used in the lower layer. 
The second is a more complex two-object transfer task that was not included in the primitive set, in order to verify whether new tasks can be generated by reusing and recombining existing motion primitives without introducing additional task-specific primitives. A schematic diagram of the two tasks is shown in Figure~\ref{fig:Setup}(\subref{fig:taskSetting}). To evaluate adaptability to environmental changes, the objects used during execution differed from those used during demonstration in terms of shape and stiffness.
All demonstration data were collected using the bilateral control method.

\subsection{Model Preparation}
In addition to the proposed methods—the learning-based, sampling-based, and playback-based proportion models—a standard hierarchical model was used as a baseline for comparison. In the baseline model, both the upper and lower layers are trained on the target task, whereas in the proposed methods, the upper layer learns the target task and the lower layers learn 50 motion primitives. To improve robustness and diversity, the number of lower-layer outputs was augmented tenfold by adding noise, resulting in a total of 500 samples. However, in the learning-based proportion model, training with 50 primitives requires the upper layer to learn all possible combinations among them, which exceeds its learning capacity. Therefore, the number of motion primitives was reduced to 30 for the learning-based proportion model.

The hyperparameters summarized in Table~\ref{tab:hyperparameter} were empirically selected to ensure stable training and reliable motion generation. All models shared the same hyperparameter settings. Although they were not exhaustively optimized, the final configuration achieved consistent and reproducible performance in robotic motion execution.
%
\begin{table}[t]
\centering
\caption{Hyperparameters for the models used in validation experiment.}
\renewcommand{\arraystretch}{1.1}
\resizebox{1.\linewidth}{!}{
    \begin{tabular}{|c|c|c|c|c|c|}
    \hline
    \textbf{Model role} & \textbf{Architecture} & \textbf{Layer} & \textbf{Neuron} & \textbf{Learning rate} & \textbf{Batch size} \\ \hline
    Upper layer & LSTM & 10 & 80  & $5 \times 10^{-4}$ & 16 \\ \hline
    Lower layer & MLP  & 8  & 200 & $5 \times 10^{-4}$ & 16 \\ \hline
    L to F      & MLP  & 10 & 80  & $5 \times 10^{-4}$ & 16 \\ \hline
    \end{tabular}}
\label{tab:hyperparameter}
\end{table}
%
\subsection{Experimental results}
\begin{table}[t]
\centering
\caption{Comparison of success rates across models.}
\renewcommand{\arraystretch}{1.}
\resizebox{.8\linewidth}{!}{
    \begin{tabular}{|c|c|c|}
    \hline
    \multirow{2}{*}{\textbf{Model}} & \multicolumn{2}{c|}{\textbf{Task success rate (\%)}} \\ \cline{2-3}
    & \textbf{Right-to-left} & \textbf{Two-object} \\ \hline
    Baseline       & 90 & 60 \\ \hline
    Learning-based & 70 & 40 \\ \hline
    Sampling-based & 100 & 70 \\ \hline
    Playback-based & 100 & 90 \\ \hline
    \end{tabular}}
    \label{tab:success-rate}
\end{table}
Both validation tasks were performed 10 times to evaluate the success rates. 
Although the number of trials is limited, consistent trends were observed across repetitions, indicating that the performance differences among models are reliable within the scope of the presented experiments. The results are summarized in Table~\ref{tab:success-rate}.

First, the baseline model achieved a success rate of 90\% in the right-to-left task but only 60\% in the two-object task, indicating that the latter task was more challenging. Two primary failure modes were observed: (1) failure to properly grasp the second object, and (2) premature termination after completing only the first pick.

Second, the learning-based proportion model showed lower success rates of 70\% and 40\% in the two tasks, respectively. 
This is attributed to the reliance on fully learned proportion prediction under uncertainty in both upper-layer planning and lower-layer motion generation, which can lead to inaccurate combinations.  
Nevertheless, the model still demonstrated the ability to generate a task not included in the primitive set in the two-object task.

In contrast, the sampling-based and playback-based proportion models performed with outstanding motion quality and achieved higher success rates. Both models achieved 100\% success in the right-to-left task, and success rates of 70\% and 90\% in the two-object task, respectively, outperforming the baseline. Their primary cause of failure was the premature termination after completing the first pick. Compared with the baseline, both models generated more stable motions and properly grasped the objects. However, the playback-based model exhibited motion-copying characteristics by executing actions synchronized with the reference motion data rather than adapting to environmental conditions or situational changes~\cite{Yokokura2008}.

It should also be noted that the proposed models were trained using a larger amount of demonstration data than the baseline, because multiple lower-layer models were trained on different task datasets. This difference may have contributed to the improved performance.

According to the above results, the sampling-based and playback-based proportion models successfully generated the complex task by reusing motion primitives. 
However, deviations were observed for the second object and the placement positions, with errors of approximately 10\,cm and 3\,cm, respectively. These deviations are attributed to the corresponding target positions lying outside the training range of the motion primitives.
Increasing the diversity and spatial range of motion primitives is expected to mitigate this limitation in future work.

Finally, the sampling-based and playback-based variants share the same lower-layer computation, and their effective update periods were comparable, operating at approximately 2.2\,ms on average, which is sufficient for real-time execution on the physical robot.
\section{Conclusions}
\label{sec5}
This study proposed an extended motion generation model based on bilateral control-based IL, which decomposes complex motions into motion primitives and accomplishes diverse tasks through their weighted combination. In the pick-and-place experiment, a complex motion not included in the primitive set was successfully generated using these primitives learned from other tasks, with task-level adaptation performed only at the upper layer. Furthermore, the lower-layer models can be easily modified and shared across different tasks, demonstrating the flexibility and reusability of the proposed architecture while reducing the overall learning cost.

Among the three proposed models, the learning-based proportion model showed lower stability when handling a large number of motion primitives, resulting in reduced accuracy. In contrast, the sampling-based proportion model achieved more stable and adaptable motion generation by learning a motion plan and the sampling-based combination. The playback-based proportion model demonstrated the fastest adaptability, requiring no upper-layer training and reproducing complex motions with high consistency.

Overall, the results confirm the effectiveness of integrating hierarchical structures with bilateral control for efficient and adaptive motion generation.
It should be noted that the proposed models were trained with a larger set of demonstration data, as the lower-layer models learned multiple tasks, which may have partially contributed to the improved performance.
Future work will focus on improving positional accuracy and enhancing generalization by introducing a world model~\cite{Ha2018} or a large-scale model~\cite{Kim2024} into the upper layer, enabling more adaptive and flexible robotic behaviors.
\balance
\bibliographystyle{IEEEtran}
\bibliography{ref}

\end{document}